\documentclass[11pt]{article}
\usepackage{eacl2017}
\usepackage{url}
\usepackage{epsfig} 
\usepackage[algo2e,ruled]{algorithm2e}
\usepackage{graphicx} 
\usepackage{subfigure}
\usepackage{hyperref,multirow,float}
\usepackage{xspace}
\usepackage{amsmath}
\usepackage{amssymb}
\usepackage{steinmetz}

\usepackage{times}
\usepackage{latexsym}
\usepackage{stmaryrd}

\graphicspath{ {images/} }

\eaclfinalcopy 

\title{Learning Semantically Coherent and Reusable Kernels in Convolution Neural Nets for Sentence Classification}
\author{Madhusudan Lakshmana \and Sundararajan Sellamanickam \\ Microsoft Research \\ Bangalore, India 
\AND Shirish Shevade \\ Indian Institute of Science, Bangalore, India \AND Keerthi Selvaraj \\ CISL, Microsoft, Sunnyvale, CA}

\date{}

\begin{document}
\maketitle
\begin{abstract}
The purpose of this work is to empirically study desirable properties such as semantic coherence, attention mechanism and kernel reusability in Convolution Neural Networks (CNNs) for learning sentence classification tasks. 
We propose to learn semantically coherent kernels using clustering scheme combined with Word2Vec representation and domain knowledge such as SentiWordNet. We also suggest a technique to visualize attention mechanism of CNNs. These ideas are useful for decision explanation purpose. Reusable property enables kernels learned on one problem to be used in another problem. This helps in efficient learning as only a few additional domain specific kernels may have to be learned. Experimental results demonstrate the
usefulness of our approach. The performance of the proposed approach, which uses
semantic and re-usability properties,
is close to that of the state-of-the-art approaches on many real-world datasets.
\end{abstract}

\section{Introduction}

In recent years, convolutional neural networks (CNNs) have proved to be very
effective in achieving state-of-the-art results for text-centric tasks~\cite{kalchbrenner2014convolutional,kim2014convolutional,hu2014convolutional}. Our focus is on learning sentence classification tasks. In a sentence classification task, the goal is to predict class label information for one or more sentences. Examples of such tasks include classifying sentiments (e.g., good or bad) and identifying question types. However, barring some limited work, there is not much discussion or empirical evidence provided on the functional behavior of the learned kernels (\textit{aka} filters) and other properties such as temporal invariance and attention mechanism capabilities. Furthermore, we are not aware of any work that provides enough evidence for the existence of properties such as semantic coherence and reusability of learned kernels. Our goal is to learn CNNs having semantically coherent kernels that are helpful in explaining the decision of the model. For the purpose of illustration, we use the CNN architecture proposed in \cite{kim2014convolutional}; however, the core ideas presented here are applicable to several other network architectures as well.   

\begin{table*}
\centering
\small
\scalebox{0.8}{
\begin{tabular}{p{4cm}p{5cm}p{5cm}}
\hline 
3 - Gram & 4-Gram & 5-Gram \\
 	 \hline 
  	 \hline 
the nagging suspicion& albeit \textbf{depressing} view of & picture postcard
perfect , too \\
 	 \hline
\textbf{unattractive} , unbearably& strangely \textbf{tempting} bouquet of &  as
the \textbf{worst} and only\\
 	 \hline 
\textbf{attractive} holiday contraption& deceptively buoyant until it & of the
best rock documentaries\\
 	 \hline 
the rich promise& sweet without relying on & simply the \textbf{best} family
film\\
 	 \hline 
're old enough& sincere but dramatically conflicted & what evans had , lost\\
 	 \hline 
\end{tabular}
}
\caption{The top scoring k-grams that match learned kernels obtained from the CNN architecture (Kim, 2014) for the \textit{Movie Review (MR)} dataset. Clearly, the k-grams are not semantically coherent.}
\end{table*}

In any classification problem, it is often important to reason out the decision made by the model. For example, we gain confidence about the model if we know that the right phrases or parts of the sentence were used by the model to predict the sentiment (e.g., \textit{positivity} in phrases such as \textit{liked such movies}). In CNNs, learning semantically coherent kernels helps in reasoning as we can identify the kernels that contribute to the decision. A kernel is semantically coherent when it fires for a collection of $k$-grams that have similar meaning (e.g., \textit{liked such movies, loved this film}). Table 1 gives an illustration of $k$-grams that fire for a few kernels in the CNN model obtained from the CNN architecture (Kim, 2014). It is clearly seen that the learned kernels are not semantically coherent. Although the learned CNN model gives good performance, it becomes difficult to explain the decision without semantic coherence. 

To address this problem, we propose to learn semantically coherent kernels. Our approach involves clustering of \textit{k-grams} that occur in the sentence corpus. We use a distance function that is a weighted combination of distances in the Word2Vec~\cite{mikolov2013efficient} and SentiWordNet~\cite{baccianella2010sentiwordnet} representation spaces, resulting in meaningfully polarized clusters. This helps to enhance semantic coherence of clusters. We define a parametrized convolution kernel for each cluster and jointly learn these kernel parameters along with weights of a linear softmax classifier output layer. The set-up for imposing semantic coherence in kernels is also well suited for obtaining solution insights by identifying or visualizing words in a sentence as used by kernels to make the decision. We propose a scoring scheme that scores each word using the \textit{max pooled} output scores of kernels and, use a simple and effective visualization technique to highlight the words in the sentence that express the sentiment; this helps to visualize the attentive capability of CNNs. 

Suppose we assume that the learned kernels have semantic coherence. Then, one question that naturally arises is: \textit{can we reuse the learned kernels from one application (or dataset, e.g., \textit{Movie Review (MR)}) to another similar application (e.g., \textit{Internet Movie Database (IMDB)})?} If the learned kernels have such a property then it is of significant help, as we can directly use these kernels without learning in several other similar applications. However, there may be some application specific semantics not covered by these fixed kernels. In such cases, we can improve the performance by learning a few additional filters but keeping the learned kernels from the other application fixed. This helps to reduce the training time significantly for new applications. 

\textbf{Contributions:} (1) We propose an approach to learn semantically coherent kernels and our experimental results show that, with semantically coherent kernels we can achieve performance close to state-of-the-art results. (2) We suggest a novel visualization technique to display words of prominence and demonstrate that both semantic coherence kernels and visualization technique help to reason out the decision. (3) We present the idea of reusability of learned kernels in similar applications and test on four different combinations of classification tasks. We find that the kernels learned in CNNs are reusable and significant reduction in training time is indeed possible.  

\section{Related Work}
CNNs have been very successful for image classification problems as they make use of internal structure of data such as the 2D structure of image data through convolution layers \cite{AK}. In the text domain, CNNs have been used for a variety of problems including sentence modeling, word embedding learning and sentiment analysis, by making use of the 1D structure of document data.

More relevant to the work in this paper
is the work of \cite{kim2014convolutional}, where it was demonstrated that a simple CNN with one layer of convolution on top of static pre-trained word vectors, obtained using Word2Vec \cite{mikolov2013efficient}, achieved excellent results on sentiment analysis and question classification tasks. \newcite{kim2014convolutional} also studied the use of multichannel representation and variable size filters.
\newcite{kalchbrenner2014convolutional} proposed  Dynamic CNN (DCNN) that alternated wide convolutional layers with dynamic $k$-max pooling to model sentences.
\newcite{Yin2} proposed Multichannel Variable-size CNN (MVCNN) architecture for sentence classification. It combines different versions of word embeddings and variable sized filters in a CNN with multiple layers of convolution.

\newcite{le2014distributed} proposed an unsupervised framework that learns continuous distributed vector representations for variable-length pieces of texts, such as sentences or paragraphs.  The vector representations are learned to predict the surrounding context words sampled from the paragraph.
The
focus is to learn paragraph vectors and task specific features are not taken into account.

\newcite{genCNN} proposed an architecture, genCNN, to predict a word sequence by exploiting both long/short range dependencies present in the history of words. \newcite{Johnson} compared the performance of two types of CNNs: seq-CNN in which every text region is represented by a fixed dimensional vector, and bow-CNN, which uses bag-of-word conversion in the convolution layer. 


\newcite{Zhang} applied CNNs only on characters and demonstrated that deep CNNs do not require the knowledge of words when trained on large-scale data sets. However, capturing semantic information using character-level CNNs is difficult.
\newcite{Santos} designed a Character to Sentence CNN (CharSCNN) that jointly uses character-level, word-level and sentence-level representations to perform sentiment analysis using two convolutional layers.

\begin{figure}
\includegraphics[scale=0.20]{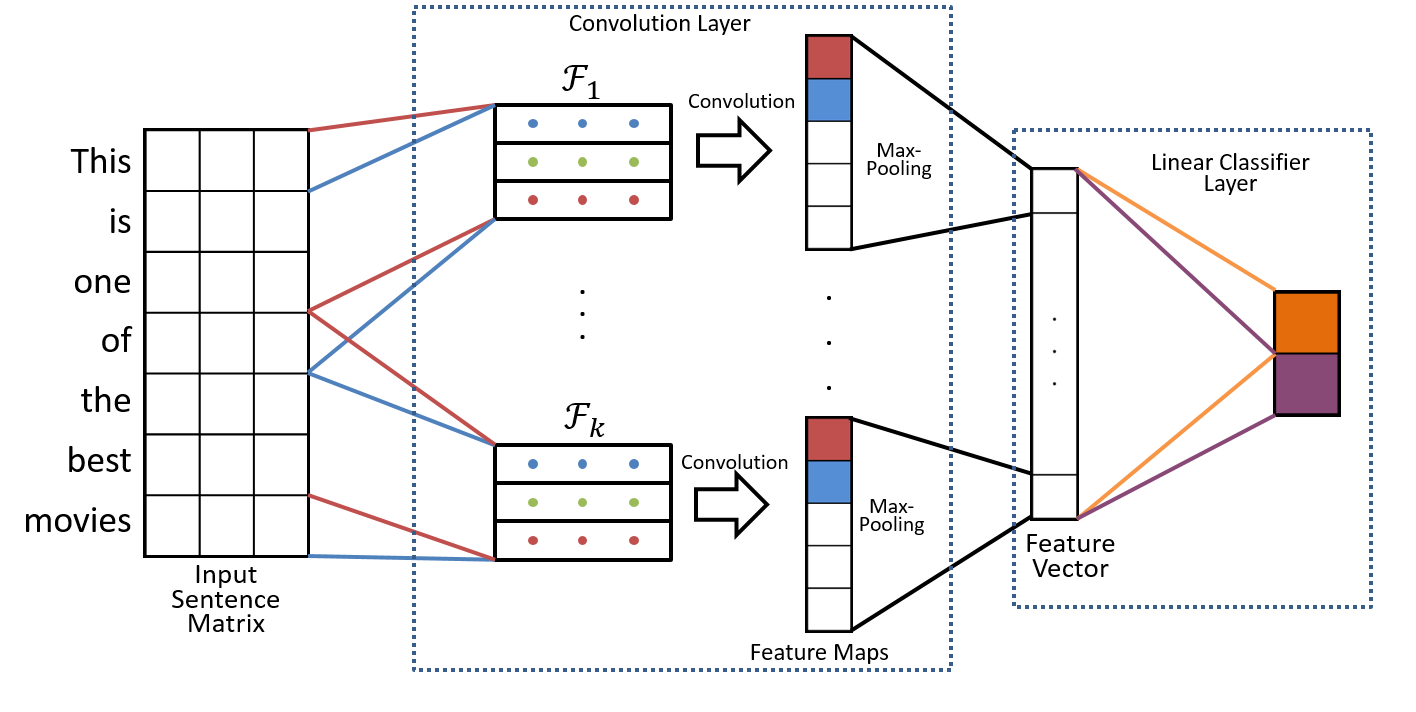}
\caption{CNN architecture from Kim (2014) used in our experiments.}
\end{figure}


\section{Learning Semantically Coherent Kernels, Visualizing Kernel Outputs and Reusable Kernels} 
In this section, we first briefly describe the CNN architecture used in this work. Then, present the notion of semantic coherence and make some observations from analyzing the filters on benchmark datasets. This is followed by our ideas to learn semantically coherent kernels and visualize the filter outputs, both aimed at helping to reason out the decision. 

\textbf{CNN Architecture.} We use the CNN architecture proposed in \cite{kim2014convolutional} (see Figure 1). This architecture has a convolution layer followed by a linear classifier layer. Let ${\bf x}_i \in {\mathcal R}^d$ denote a $d$-dimensional representation (e.g., Word2Vec) of the $i^{th}$ word in a sentence. We concatenate these vectors as ${\bf x}_{1:n}$ to represent the sentence. Each kernel $v_j \in {\mathcal R}^{dk}$ is convolved with a sentence to produce a single feature output $g_j$ as follows. With $k$ representing the kernel width, a feature map ${\bf f}_j$ is produced with the $i^{th}$ feature value computed as: $f_{j,i} = ReLU({\bf v}^T_j {\bf x}_{i:i+k-1})$ where $ReLU$ denotes the rectified linear unit function ($ReLU(x) = 0, x < 0$ and $ReLU(x) =  x, x \ge 0$). This is followed by a max-pooling operation to produce the feature output as $g_j = max\{f_{j,1},\cdots,f_{j,n-k+1}\}$. Thus, the convolution layer produces an $m$ dimensional feature vector ${\bf g}$ using $m$ kernels. The second layer is a linear classifier layer that computes the $c^{th}$ class probability score as: $p_c = \frac{exp(s_c)}{\sum_{{\tilde c}=1}^K exp(s_{\tilde c})}$ where $s_c = {\bf w}^T_c {\bf g}$ and ${{\bf w}_c}$ is the weight vector corresponding to the $c^{th}$ class.

We use Word2Vec representations learned from Google News data, for representing words. Each word is represented as a $300$ dimensional real valued vector. Following ~\newcite{kim2014convolutional}, we use $3$ kernel widths ($k=3,4,5$) in our experiments; note that these kernel widths correspond to convolving through $3$-grams, $4$-grams and $5$-grams respectively in a sentence. Unless otherwise specified, we use $100$ kernels for each $k$ resulting in $m=300$.  

\begin{table*}
\centering
\small
\begin{tabular}{p{3cm}p{4cm}p{5cm}p{3.5cm}}
\hline 
3 - Gram & 4-Gram & 5-Gram & 3 - Gram (not so good)\\
 	 \hline 
  	 \hline 
enjoy the film& impressive and highly & by sumptuous ocean visuals and& neurotic , and\\
 	 & entertaining & & \\
 	 \hline
enjoy this movie& year 's most intriguing & a fascinating document of an& sincere grief and\\
 	 \hline 
liked this film& out of the intriguing & a fascinating portrait of a& self important and\\
 	 \hline 
enjoying this film& the characters are intriguing & each interesting the movie is& romantic problems of\\
 	 \hline 
appreciate the film& strong and politically potent & and beautifully rendered film one& self important ,\\
 	 \hline 
\end{tabular}
\caption{The top scoring $k$-grams that match learned semantic coherent kernels for the \textsc{MR} dataset. Clearly, the first three kernels are semantically coherent. The last kernel is not as coherent as others.}
\end{table*}

\subsection{Learning Semantically Coherent Kernels}

Our goal is to design kernels where each kernel captures some semantics (e.g., \textit{liked such movies, loved this film}). We call a set of $k$-grams that have similar meaning as semantically coherent. We expect a semantically coherent kernel to produce high scores for $k$-grams having similar meaning. Table 1 gives an illustration of $k$-grams that have top-$5$ highest cosine similarity scores for a few kernels using ~\newcite{kim2014convolutional}'s network depicted in Figure 1. We see that the top-$5$ $k$-grams do not have similar meaning; thus, the learned kernel lacks semantic coherence. Therefore, it is not clear what these kernels represent and how to use high scoring filter outputs to reason out the decision.    

To learn semantically coherent kernels, we take a two-step approach. In the first step, we select a subset of $k$-grams from the sentence corpus and group them into a desired number of clusters. This clustering step helps to group $k$-grams that are semantically coherent. We associate a kernel ${\bf v}$ with each cluster as a weighted combination of Word2Vec representations of the $k$-grams that are members of the cluster. More formally, ${\bf v}_j = \sum_{l \in {\mathcal C}_j} z_{l} {\bf p}_l$ where ${\mathcal C}_j$ denote the set of indices for the $k$-grams that constitute the $j^{th}$ cluster; $z_{l}$ and ${\bf p}_l$ denote the learnable weight and concatenated Word2Vec representation associated with the $l^{th}$ $k$-gram. Thus, each kernel is parametrized and we learn the parameters of the kernels (i.e., $z_{l} \forall l $) jointly with the weights (${\bf w}_c \forall c$) of the linear classifier layer. We call this model as Weighted $k$-gram Averaging (WkA). A naive approach is to represent the kernel as the centroid of the cluster, i.e., set $z_{l} = \frac{1}{|{\mathcal C}_j|} \forall l \in {\mathcal C}_j$. As we show in the experiments section, significant performance improvement is achieved by learning the kernel parameters. We note that the WkA model is built using the same architecture (Figure 1) but our kernels ${\bf v}_j \forall j$ are constrained via the parameters $z_{l} \forall l$ to lie in the subspaces spanned by the $k$-grams in ${\mathcal C}_j \forall j$. On the other hand, the kernel parameters ${\bf v}_j \forall j$ are learned by optimizing them as free variables in Kim's model using the same architecture. Therefore, we can expect some degradation in the performance of our model; but, we learn semantically coherent kernels that are easy to use for explanation purpose. 

\textbf{Selecting $k$-grams.} The WkA model complexity is dependent on the number of $k$-grams that are used to form the kernels. Since the number of $k$-grams can be very large in the data corpus, it helps to control the model complexity by learning with a selected set of $k$-grams. We experimented with three simple selection heuristics. Details are given in Section 4 and supplemental material. 

\textbf{Clustering using Domain Knowledge.} To perform clustering, we need to define a distance function.  Since we represent a $k$-gram using the Word2Vec representation that captures distributional semantics using contextual information in ${\mathcal R}^d$, Euclidean distance is a good distance function to use. We discover the clusters using the $K$-means algorithm. However, visual inspection showed that the quality of clusters was not good. The main reason is that some words with opposite meanings get similar representations (due to similar contexts in which they occur while learning Word2Vec representation). Some examples with cosine similarity score are: \textit{(attractive,unattractive,0.72)}, \textit{(good,bad,0.71)}, \textit{(able,unable,0.68)}, \textit{(bright,dim,0.59)} and \textit{(worst,best,0.58)}. This is not desirable in applications such as sentiment classification where $k$-grams with opposite sentiments are not semantically coherent. Therefore, it is important to form sentiment polarized clusters in such applications; that is, $k$-grams expressing the same sentiment and semantic should occur together in every cluster. To form sentiment polarized clusters, we need an additional representation of $k$-grams that can capture the sentiment. For this purpose, we bring in domain knowledge via SentiWordNet knowledge base~\cite{esuli2006sentiwordnet,baccianella2010sentiwordnet}. Using this knowledge base, we assign a sentiment score for each word as explained below.

\textbf{SentiWordNet Representation.} SentiWordNet gives a 2-tuple of positive and negative scores for each sense of a word. There are several ways in which we can assign a SentiWordNet score for a word in a sentence. The best way is to find the sense of the word and use the corresponding 2-tuple. Simpler techniques are to aggregate the 2-tuples by averaging or using the maximum element-wise score in the tuples. In our experiments, we found that the \textit{maximum} aggregation technique works well. Thus, the SentiWordNet representation of a $k$-gram is a $2k$ dimensional vector.


{\bf Forming Sentiment Polarized Clusters.}  We concatenate the Word2Vec and SentiWordNet representations. Note that these representations capture the semantic information derived from context seen in a large corpus and sentiment information derived from task specific data corpus respectively. Given the joint representation, we modify the distance function as a weighted combination of distance functions in the Word2Vec and SentiWordNet representation spaces. We set these weights by manually inspecting the quality of clusters. Table 2 shows a few kernels with top scoring $k$-grams obtained with our learned semantically coherent kernels for the \textsc{MR} (Movie Review) dataset. We see that most of the kernels are semantically coherent. Though the $k$-grams in the last column are noisy, we can improve the semantic coherent quality of kernels using better distance functions and optimizing the weights by treating them as hyperparameters.  

While Table 2 is useful to qualitatively assess semantic coherence by visual inspection, we could make use of the weighted distance function (described above) to define a computable semantic coherent score for each cluster as follows. We compute a normalized average score ($G_j$) of weighted distances of all pairs in the $j^{th}$ cluster ${\mathcal C}_j$. Then, we define the semantic coherence score as: $S_j = 1-G_j$; the normalization is done so that the semantic coherence score lies in the interval $[0,1]$. Higher values of $S_j$ indicate stronger semantic coherence. To reduce computational cost, we computed $S_j$ for each filter using only top scoring $50$ $k$-grams. Figure 2(a) shows the semantic coherence scores of $300$ learned filters on the \textsc{MR} dataset. We see that the weighted $k$-gram model has significantly higher mass towards the right as compared to the CNN-Static model~\cite{kim2014convolutional}. For example, the WkA model has $48\%$ of filters with score more than $0.6$; this is significantly higher compared to $7\%$ filters in the CNN-Static model. 

Optimizing hyperparameters using averaged semantic coherence score over the clusters as the objective function is left as a future research work.     

\begin{figure*}
\subfigure[]{\includegraphics[scale=0.3]{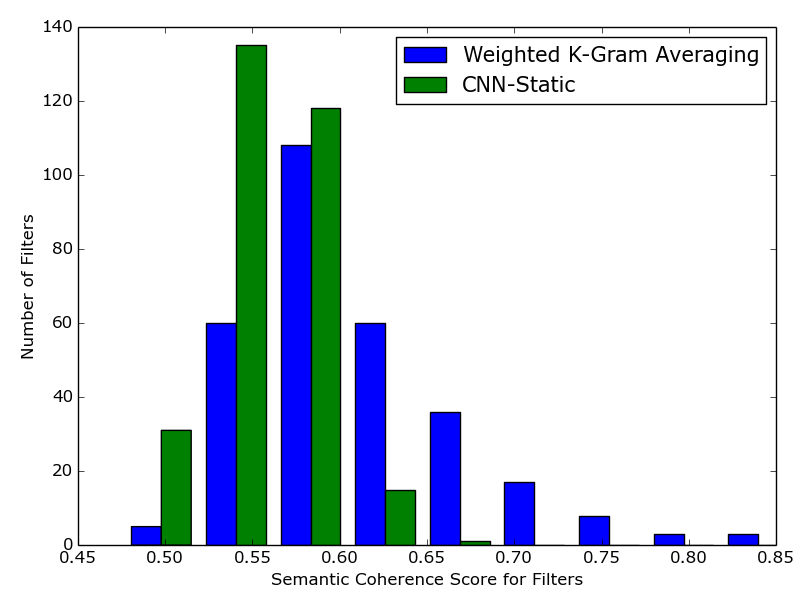}}
\subfigure[]{\includegraphics[scale=0.45]{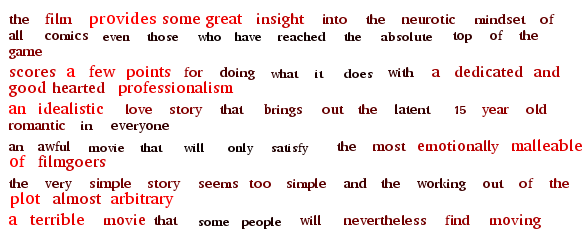}}
\caption{Figure (a) shows the histogram of semantic coherence scores of $300$ learned filters from the WkA and CNN-Static (Kim, 2014) models on the \textsc{MR} dataset. 
The WkA model exhibits higher semantic coherence. Figure (b) illustrates our visualization technique for positive (top $3$) and negative sentences in a sentiment classification task.} 
\end{figure*}

\subsection{Visualizing Kernel Outputs} 
We propose a simple but effective technique to visualize words (in a sentence) that are discovered as important by kernels in making the decision. In Figure 2(b), we present a few sentences with words marked with a graded color map and font sizes; the marked words with red/dark red colors with larger font sizes are identified as important using our approach by the learned semantically coherent kernels. 

We generated these marked sentences as follows. For a given sentence, we identify the $k$-gram that fires as the \textit{max pooled} output of each kernel. Then, we associate respective weighted kernel output score for each word in the selected $k$-gram for all kernels; here, the weight can be set to $1$ or as the linear classifier feature weight depending on what we would like to visualize. Finally, we sum the scores for each word, as the same word can be part of multiple selected $k$-grams and normalize the scores to the range $[0,255]$. The normalized scores are used as intensity values in a graded color map. For example, a zero intensity value represents \textit{black} color and the highest value of $255$ corresponds to \textit{dark red} color. Furthermore, it helps to use higher font sizes for ease of visualization. For example, we mapped $5$ increasing font sizes to $5$ equally spaced intensity range intervals, for generating Figure 2(b). We see that several highlighted important $k$-grams nicely represent the sentiments that help to reason out the decision and it also illustrates the attention capability of CNNs. 

\subsection{Reusable Kernels} We call a kernel reusable when a kernel learned in one application (or dataset, e.g., \textsc{MR}) serves as a useful kernel in similar applications (e.g., \textsc{IMDB}). We expect this to happen in CNNs when $k$-gram models are used and similar $k$-grams appear in similar applications (e.g., across movie review datasets, across electronic product review datasets). In particular, since we learn semantically coherent kernels, we expect this property to hold as these kernels represent distinct semantic notions as seen in Table 2. There are several ways to use learned kernels on a new dataset. A simple baseline is to use them as fixed kernels and learn only the classifier layer outputs. Another way is to adjust the weights of $k$-grams in each kernel with weight regularization using previously learned weights. We can extend further by adding a few more kernels and learn them either with fixed or weight regularized reusable kernels. 

In our experiments, we used fixed kernels and learning with additional filters. The need for additional kernels arises because some domain specific $k$-grams will often be present and significant improvement in performance can be achieved by using additional filters to cover these $k$-grams. One key advantage of using reusable kernels is that we can achieve significant reduction in training time on new applications as we need to learn only smaller number of parameters. As we show in the next section, $20-50$ times speed-up is possible on real-world problems.

\section{Experimental Evaluation}
In this section, we demonstrate the efficacy of learning semantically coherent kernels by comparing the performance with a few baselines and the CNN-Static model~\cite{kim2014convolutional} (Figure 1). As emphasized earlier, our core ideas can be easily extended and applied in other sophisticated CNN models (e.g., multichannel~\cite{kim2014convolutional} and learning Word2Vec representations). We also demonstrate through several examples that the learned kernels in CNNs can be reused in similar applications and significant reduction in training time can be achieved. Overall, we are able to achieve performance close to the state-of-the-art methods but with semantic and reusable properties.      

\begin{table} 
          \begin{tabular}{|c|c|c|c|} 
     \hline  
     Dataset & Train  &  Validation  & Test \\ 
     & Split Size & Split Size & Split Size\\ 
     \hline \hline 
     MR & 10662 & 10-CV & - \\ 
      \hline  
      SST-1 & 8544 & 1101 & 2210\\ 
      \hline  
    SST-2 & 6920 & 872 &1821\\ 
     \hline  
    SUBJ & 10000 & 10-CV & -\\ 
     \hline  
    IMDB & 22500 & 2500 &25000\\ 
     \hline  
      \end{tabular} 
      
    \caption{\small{Statistics of Datasets. (CV: Cross Validation)}} 
\end{table}

\begin{table*}
\centering
\begin{tabular}{|l|l|l|l|l|l|}
\hline 
\textbf{Model} & \textbf{MR} & \textbf{IMDB} & \textbf{SUBJ} &
\textbf{SST-1} & \textbf{SST-2}\\
\hline
Simple Word2Vec Averaging & 76.76 & 82.09 & 88.03 & 39.28 & 78.36\\
\hline
Weighted Word2Vec Averaging & 77.97 & 88.39 & 91.55 & 44.34 & 82.98 \\
\hline
\hline
Simple $k$-gram Averaging (SkA) & 61.67 & 58.02 & 70.65 & 32.21 & 61.89\\
\hline 
Weighted $k$-gram Averaging (WkA) & 78.65 & 88.76 & 91.95 & 45.97 & 83.25 \\
\hline
WkA + Parse Tree & 78.68 & 89.24 & 92.17 & 46.33 & 83.86 \\
\hline
WkA + POS-Tag & 79.04 & 89.24 & 92.02 & 45.75 & 83.47\\
\hline
WkA + 10\% flexible filters (FF) & 79.75 & 89.45 & 92.51 & 45.84 & 83.69\\
\hline
WkA + 25\% flexible filters (FF) & 80.02 & 90.16 & 92.68 & 46.11 & 84.29\\ 
\hline
CNN-Static (CNN-S) \cite{kim2014convolutional} & {\it 81.0} & 90.85$^{*}$ & {\it 93.0} & {\it 45.5} & {\it 86.8}\\
\hline
DCNN \cite{kalchbrenner2014convolutional} & - & - & - & {\it 48.5} & {\it 86.8}\\
\hline
MV-RNN \cite{socher2012recursive} & {\it 79.0} & - & - & {\it 44.4} & {\it 82.9}\\
\hline
  \end{tabular}
\caption{CNN-Static refers to a CNN model with Word2Vec representation. DCNN refers to dynamic CNN with k-max pooling. MV-RNN refers to Matrix-Vector Recursive Neural Network with parse
trees. The numbers in italics are as reported in the respective papers. The number with $*$ was obtained by executing the code quoted in (Kim, 2014) where the IMDB dataset result was not reported.} 
\end{table*}

\subsection{Experimental Setup}
{\bf Datasets.} We conducted a comprehensive set of our experiments on $5$ popular benchmark datasets used for sentence classification tasks~\cite{kim2014convolutional}. They are: \textsc{MR}, \textsc{IMDB}, \textsc{SST-1}, \textsc{SST-2} and \textsc{SUBJ} (Subjectivity). The first $4$ tasks (datasets) are sentiment classification tasks and all are binary classification tasks except \textsc{SST-1} (which has fine-grained sentiment labels with $5$ classes). The \textsc{SUBJ} dataset is again a binary classification task where sentences are labeled as \textit{Subjective} or \textit{Objective}. The statistics of the datasets are given in Table 3. More details can be found in the supplemental material. 

{\bf Models.} We compare the performance of several models. All the methods differ in the sentence model (representation) they form, i.e., the feature input vector that forms the input to the classifier layer. We can categorize these methods into three categories. The first category of models aggregate the Word2Vec representations of words in a sentence and they do not use convolution filters. We have two baselines in this category. The first baseline uses a sentence model that averages the Word2Vec representations with equal weights. In the second baseline, we assign a weight for each word in the vocabulary and form the sentence representation as a weighted combination; we learn these weights jointly with the classifier layer weights. The models learned using these methods are referred as \textit{Simple Word2Vec Averaging} and \textit{Weighted Word2Vec Averaging} in Table 4. 

The second category of models uses convolution filter (kernel) representation obtained using $k$-gram clusters; here again, we have simple averaging and weighted averaging of $k$-gram Word2Vec representations to form the kernel representation. These methods are referred as \textit{Simple $k$-gram Averaging} (SkA)  and \textit{Weighted $k$-gram Averaging} (WkA) in Table 4. The number of $k$-grams can be extremely large. For example, the number of unique $3$-grams in \textsc{MR} and $\textsc{IMDB}$ datasets is given by $169000$ and $6.5$ million respectively. Therefore, as discussed earlier, we experimented with three different heuristics in order to control the complexity. In the first heuristic, we shortlist the $k$-grams (e.g., around $50000 - 100000$ for each $k$) using sentiment information available in each $k$-gram; we used a dictionary of positive and negative sentiment words and, if a $k$-gram does not contain any sentiment word from this dictionary, we drop it. We did not use any shortlisting using sentiment dictionary in the $\textsc{SUBJ}$ dataset since this is not a sentiment classification problem; but, we sampled $100000$ $k$-grams for each $k$.  Results obtained using the first heuristic are referred as the row WkA in Table 4. With the goal of improving the coverage over $k$-grams, we also identified additional $k$-grams that are obtained as outputs of a syntactic tree parser. As the third heuristic, we selected $k$-grams that contain at least one word belonging to the POS tag categories of verbs, adjectives and their derivatives. The results obtained using these heuristics are reported as: WkA + Parse Tree and WkA + POS-Tag respectively.   


The third category of models are some of the CNN~\cite{kim2014convolutional,kalchbrenner2014convolutional} and RNN models~\cite{socher2012recursive} reported in the literature. We emphasize that our intention is not to get the best performance using complex network architectures; but, to learn network models where the learned kernels are semantically coherent which helps to reason out through inspection of fired features and visualization technique. \\
{\bf Training and Hyper-parameter Settings.} We used $L_1$ regularized negative log likelihood function (i.e., $\lambda (|{\bf w}| + |{\bf z}|) - \frac{1}{n} \sum_{i=1}^n \log p(y_i|{\bf X}_i;{\bf w},{\bf z})$) as the objective function to learn the model parameters. Here, ${\bf X}_i$ and $y_i$ denote the input sentence representation and class label information respectively for the $i^{th}$ example, with the class probabilities defined using the softmax function explained earlier; ${\bf w}$, ${\bf z}$ and $n$ denote the linear classifier layer weights (collated over the classes), kernel parameters (collated over all the clusters) and number of examples respectively. Note that the kernel parameters ${\bf z}$ are nothing but the free variables ${\bf v}$ when we trained the CNN-S model. We trained the models for different regularization constants and many passes ($50-100$) over the training set using mini-batch with \textit{AdaDelta} learning rate updates~\cite{zeiler2012adadelta} and Dropout~\cite{srivastava2014dropout} of 0.5. We chose the model that gives the best validation accuracy and report the test set accuracy for this model. To compute the distance function for clustering, we set the weights for the distance functions after several experimentation and visualizing the quality of clusters. We used $100$ kernels each for $k$-grams with $k=3,4,5$~\cite{kim2014convolutional}. More details can be found in the supplemental material. 

\subsection{Experimental Results} 
{\bf Comparison of Models.} Table 4 gives the test accuracy results of the various models described in Section 4.1 on $5$ benchmark datasets. It is interesting to see that weighted averaging of Word2Vec representation gives reasonable performance. Learning the weights for the sentence models significantly improves the performance; this can be seen by comparing pairs of results in (first,second) and (third, fourth) rows for Word2Vec and $k$-gram based models respectively. Recall that we learn semantically coherent kernels in the second category of models. and we see that the WkA model gives similar performance compared to the flexible but non-interpretable CNN-S model. 
As we can see, the performance of the WkA model is quite good even with such a limited set of $k$-grams. And, the performance improves significantly on several datasets as we bring in additional $k$-grams using syntactic parse tree and POS tag based information to form the clusters. We see that there is still some performance gap with the CNN-S model. As explained earlier, one reason is that our kernel parameters are constrained to be part of the subspaces spanned by the clusters, as opposed to treating them as free variables in the CNN-S model. This can be addressed, if needed, by adding a few filters (e.g., $10\%$ or $25\%$ of the total number of semantically coherent kernels) and learn these filters jointly with semantically coherent filters. These models are referred as WkA with respective percentage of flexible filters added. As we can see, there is a clear trend in performance improvement as more filters are added and the performance gap reduces significantly with the CNN-S model. But, we lose some reasoning capability, as we would not be able to explain the reason when any of these added filters fire in making a decision.
Note that we can trade-off between interpretability and improved performance accuracy, by controlling the percentage of flexible filters. When reasoning out is an important requirement with limited accuracy loss (e.g., $1$-$2\%$ as seen from the table), the approach of learning semantically coherent kernels can be significantly useful. Furthermore, our approach nicely discovers the important words as highlighted with the visualization technique demonstrated earlier in Figure 2. Overall, we see that the approach of learning semantically coherent kernels is quite effective. 

{\bf Results from Reusable Kernel Experiment.} We conducted these experiments on $4$ datasets \textsc{MR}, \textsc{SST-1}, \textsc{SST-2} and \textsc{IMDB}. Since the first three datasets share common sentences, we reused kernels learned from \textsc{IMDB} for these datasets. As another experiment, we reused kernels learned from \textsc{MR} on \textsc{IMDB}. From Table 5, we see that decent accuracy is achievable by using fixed kernels from WkA and CNN-S models. The reason behind the observed significant performance difference of these models with fixed kernels is that the CNN-S model uses all the $k$-grams to form the kernel. As we can see, the performance gap is significantly reduced by adding just $10\%$ of additional kernels; note that we added the same number of kernels to CNN-S model for fair comparison. But, the improvement achieved by our model is more ($4.2\%$) as the $k$-gram coverage is improved significantly; it is just $1.6\%$ with the CNN-S model as it has covered a larger fraction already. Overall, we see that the kernels learned in CNNs are indeed reusable with both WkA and CNN-S models. 

We measured the training time taken with full training (i.e., no reusable kernels) of our WkA model and with fixed kernels on the \textsc{SST-1} dataset. While the full training takes nearly $2$ hours, it just takes $2$ minutes with fixed kernels. Note that only the classifier layer needs to be learned with fixed reusable kernels. Adding $10\%$ kernels increased the training time to approximately $4$ minutes. Thus, we see an order of magnitude improvement in training time while achieving similar performance.  

\begin{table}
\begin{small}
\centering
\begin{tabular}{|l|l|l|l|l|}
\hline 
\textbf{Model} & \textbf{MR} & \textbf{IMDB} &
\textbf{SST-1} & \textbf{SST-2}\\
\hline
Fixed (WkA) & 73.78 & 82.50 & 41.76 & 80.67 \\
\hline
Fixed (CNN-S) & 77.72 & 85.78 & 44.43 & 82.81 \\
\hline
Fixed (WkA)  & 78.77 & 89.17 & 43.30 & 84.24 \\
+10\% FF  &  & & & \\
\hline
Fixed (CNN-S)& 79.73 & 89.54 & 43.57 & 84.24\\
+10\% FF  &  & & & \\
\hline
\end{tabular}
 \\
\caption{Reusable Kernel Experiment Results.} 
\end{small}
\end{table}
\section{Conclusion} 
In this work, we proposed to learn semantically coherent kernels using clustering scheme combined with Word2Vec representation and domain knowledge such as SentiWordNet. We suggested an effective technique to visualize words discovered by kernels. Semantically coherent kernels and identifying prominent words help to reason out the decision. We introduced kernel reusability and showed that kernels learned in one application are useful in similar applications, achieving close to state-of-the-art performance but with reduced training time. 

\nocite{bengio2003neural}
\nocite{duchi2011adaptive}
\nocite{wang2012baselines}
\nocite{pang2004sentimental}
\nocite{rong2014word2vec}
\nocite{brown1992class}
\nocite{Turian}
\nocite{Luo}
\nocite{socher2013recursive}
\nocite{ABC-CNN}
\nocite{Yin}
\nocite{Tang}
\nocite{mou2015discriminative}
\nocite{zhang2016dependency}
\nocite{kumar2015ask}
\nocite{zhao2016attention}
\nocite{tai2015improved}
\nocite{wan2013regularization}


\bibliography{eacl2017}

\begin{thebibliography}{}

\bibitem[\protect\citename{Baccianella \bgroup et al.\egroup
  }2010]{baccianella2010sentiwordnet}
Stefano Baccianella, Andrea Esuli, and Fabrizio Sebastiani.
\newblock 2010.
\newblock Sentiwordnet 3.0: An enhanced lexical resource for sentiment analysis
  and opinion mining.
\newblock In Nicoletta Calzolari~(Conference Chair), Khalid Choukri, Bente
  Maegaard, Joseph Mariani, Jan Odijk, Stelios Piperidis, Mike Rosner, and
  Daniel Tapias, editors, {\em Proceedings of the Seventh conference on
  International Language Resources and Evaluation (LREC'10)}, Valletta, Malta,
  may. European Language Resources Association (ELRA).

\bibitem[\protect\citename{Bengio \bgroup et al.\egroup
  }2003]{bengio2003neural}
Yoshua Bengio, R{\'e}jean Ducharme, Pascal Vincent, and Christian Janvin.
\newblock 2003.
\newblock A neural probabilistic language model.
\newblock {\em The Journal of Machine Learning Research}, 3:1137--1155.

\bibitem[\protect\citename{Brown \bgroup et al.\egroup }1992]{brown1992class}
Peter~F Brown, Peter~V Desouza, Robert~L Mercer, Vincent J~Della Pietra, and
  Jenifer~C Lai.
\newblock 1992.
\newblock Class-based n-gram models of natural language.
\newblock {\em Computational linguistics}, 18(4):467--479.

\bibitem[\protect\citename{Chen \bgroup et al.\egroup }2015]{ABC-CNN}
Kan Chen, Jiang Wang, Liang{-}Chieh Chen, Haoyuan Gao, Wei Xu, and Ram Nevatia.
\newblock 2015.
\newblock {ABC-CNN:} an attention based convolutional neural network for visual
  question answering.
\newblock {\em CoRR}, abs/1511.05960.

\bibitem[\protect\citename{dos Santos and Gatti}2014]{Santos}
Cicero dos Santos and Maira Gatti.
\newblock 2014.
\newblock Deep convolutional neural networks for sentiment analysis of short
  texts.
\newblock In {\em Proceedings of COLING 2014, the 25th International Conference
  on Computational Linguistics: Technical Papers}, pages 69--78, Dublin,
  Ireland, August. Dublin City University and Association for Computational
  Linguistics.

\bibitem[\protect\citename{Duchi \bgroup et al.\egroup
  }2011]{duchi2011adaptive}
John Duchi, Elad Hazan, and Yoram Singer.
\newblock 2011.
\newblock Adaptive subgradient methods for online learning and stochastic
  optimization.
\newblock {\em The Journal of Machine Learning Research}, 12:2121--2159.

\bibitem[\protect\citename{Esuli and Sebastiani}2006]{esuli2006sentiwordnet}
Andrea Esuli and Fabrizio Sebastiani.
\newblock 2006.
\newblock Sentiwordnet: A publicly available lexical resource for opinion
  mining.
\newblock In {\em Proceedings of LREC}, volume~6, pages 417--422. Citeseer.

\bibitem[\protect\citename{Hu \bgroup et al.\egroup }2014]{hu2014convolutional}
Baotian Hu, Zhengdong Lu, Hang Li, and Qingcai Chen.
\newblock 2014.
\newblock Convolutional neural network architectures for matching natural
  language sentences.
\newblock In {\em Advances in Neural Information Processing Systems}, pages
  2042--2050.

\bibitem[\protect\citename{Johnson and Zhang}2014]{Johnson}
Rie Johnson and Tong Zhang.
\newblock 2014.
\newblock Effective use of word order for text categorization with
  convolutional neural networks.
\newblock {\em CoRR}.

\bibitem[\protect\citename{Kalchbrenner \bgroup et al.\egroup
  }2014]{kalchbrenner2014convolutional}
Nal Kalchbrenner, Edward Grefenstette, and Phil Blunsom.
\newblock 2014.
\newblock A convolutional neural network for modelling sentences.
\newblock In {\em Proceedings of the 52nd Annual Meeting of the Association for
  Computational Linguistics (Volume 1: Long Papers)}, pages 655--665,
  Baltimore, Maryland, June. Association for Computational Linguistics.

\bibitem[\protect\citename{Kim}2014]{kim2014convolutional}
Yoon Kim.
\newblock 2014.
\newblock Convolutional neural networks for sentence classification.
\newblock In {\em Proceedings of the 2014 Conference on Empirical Methods in
  Natural Language Processing (EMNLP)}, pages 1746--1751, Doha, Qatar, October.
  Association for Computational Linguistics.

\bibitem[\protect\citename{Krizhevsky \bgroup et al.\egroup }2012]{AK}
Alex Krizhevsky, Ilya Sutskever, and Geoffrey~E Hinton.
\newblock 2012.
\newblock Imagenet classification with deep convolutional neural networks.
\newblock In {\em Advances in neural information processing systems}, pages
  1097--1105.

\bibitem[\protect\citename{Kumar \bgroup et al.\egroup }2015]{kumar2015ask}
Ankit Kumar, Ozan Irsoy, Jonathan Su, James Bradbury, Robert English, Brian
  Pierce, Peter Ondruska, Ishaan Gulrajani, and Richard Socher.
\newblock 2015.
\newblock Ask me anything: Dynamic memory networks for natural language
  processing.
\newblock {\em arXiv preprint arXiv:1506.07285}.

\bibitem[\protect\citename{Le and Mikolov}2014]{le2014distributed}
Quoc~V Le and Tomas Mikolov.
\newblock 2014.
\newblock Distributed representations of sentences and documents.
\newblock {\em In Proceedings of ICML}.

\bibitem[\protect\citename{Luo \bgroup et al.\egroup }2014]{Luo}
Yong Luo, Jian Tang, Jun Yan, Chao Xu, and Zheng Chen.
\newblock 2014.
\newblock Pre-trained multi-view word embedding using two-side neural network.
\newblock In {\em AAAI}, pages 1982--1988.

\bibitem[\protect\citename{Maas \bgroup et al.\egroup
  }2011]{maas-EtAl:2011:ACL-HLT2011}
Andrew~L. Maas, Raymond~E. Daly, Peter~T. Pham, Dan Huang, Andrew~Y. Ng, and
  Christopher Potts.
\newblock 2011.
\newblock Learning word vectors for sentiment analysis.
\newblock In {\em Proceedings of the 49th Annual Meeting of the Association for
  Computational Linguistics: Human Language Technologies}, pages 142--150,
  Portland, Oregon, USA, June. Association for Computational Linguistics.

\bibitem[\protect\citename{Mikolov \bgroup et al.\egroup
  }2013]{mikolov2013efficient}
Tomas Mikolov, Kai Chen, Greg Corrado, and Jeffrey Dean.
\newblock 2013.
\newblock Efficient estimation of word representations in vector space.
\newblock {\em ICLR Workshop, 2013}.

\bibitem[\protect\citename{Mou \bgroup et al.\egroup
  }2015]{mou2015discriminative}
Lili Mou, Hao Peng, Ge~Li, Yan Xu, Lu~Zhang, and Zhi Jin.
\newblock 2015.
\newblock Discriminative neural sentence modeling by tree-based convolution.
\newblock page 2315–2325.

\bibitem[\protect\citename{Pang and Lee}2004]{pang2004sentimental}
Bo~Pang and Lillian Lee.
\newblock 2004.
\newblock A sentimental education: Sentiment analysis using subjectivity
  summarization based on minimum cuts.
\newblock In {\em Proceedings of the 42nd annual meeting on Association for
  Computational Linguistics}, page 271. Association for Computational
  Linguistics.

\bibitem[\protect\citename{Pang and Lee}2005]{pang2005}
Bo~Pang and Lillian Lee.
\newblock 2005.
\newblock Seeing stars: Exploiting class relationships for sentiment
  categorization with respect to rating scales.
\newblock In {\em Proceedings of the 43rd Annual Meeting of the Association for
  Computational Linguistics (ACL'05)}, pages 115--124, Ann Arbor, Michigan,
  June. Association for Computational Linguistics.

\bibitem[\protect\citename{Rong}2014]{rong2014word2vec}
Xin Rong.
\newblock 2014.
\newblock word2vec parameter learning explained.
\newblock {\em arXiv preprint arXiv:1411.2738}.

\bibitem[\protect\citename{Socher \bgroup et al.\egroup
  }2012]{socher2012recursive}
Richard Socher, Brody Huval, Christopher~D. Manning, and Andrew~Y. Ng.
\newblock 2012.
\newblock Semantic compositionality through recursive matrix-vector spaces.
\newblock In {\em Proceedings of the 2012 Joint Conference on Empirical Methods
  in Natural Language Processing and Computational Natural Language Learning},
  pages 1201--1211, Jeju Island, Korea, July. Association for Computational
  Linguistics.

\bibitem[\protect\citename{Socher \bgroup et al.\egroup
  }2013]{socher2013recursive}
Richard Socher, Alex Perelygin, Jean Wu, Jason Chuang, Christopher~D. Manning,
  Andrew Ng, and Christopher Potts.
\newblock 2013.
\newblock Recursive deep models for semantic compositionality over a sentiment
  treebank.
\newblock In {\em Proceedings of the 2013 Conference on Empirical Methods in
  Natural Language Processing}, pages 1631--1642, Seattle, Washington, USA,
  October. Association for Computational Linguistics.

\bibitem[\protect\citename{Srivastava \bgroup et al.\egroup
  }2014]{srivastava2014dropout}
Nitish Srivastava, Geoffrey~E Hinton, Alex Krizhevsky, Ilya Sutskever, and
  Ruslan Salakhutdinov.
\newblock 2014.
\newblock Dropout: a simple way to prevent neural networks from overfitting.
\newblock {\em Journal of Machine Learning Research}, 15(1):1929--1958.

\bibitem[\protect\citename{Tai \bgroup et al.\egroup }2015]{tai2015improved}
Kai~Sheng Tai, Richard Socher, and Christopher~D Manning.
\newblock 2015.
\newblock Improved semantic representations from tree-structured long
  short-term memory networks.
\newblock {\em arXiv preprint arXiv:1503.00075}.

\bibitem[\protect\citename{Tang}2015]{Tang}
Duyu Tang.
\newblock 2015.
\newblock Sentiment-specific representation learning for document-level
  sentiment analysis.
\newblock In {\em Proceedings of the Eighth ACM International Conference on Web
  Search and Data Mining}, pages 447--452. ACM.

\bibitem[\protect\citename{Turian \bgroup et al.\egroup }2010]{Turian}
Joseph Turian, Lev-Arie Ratinov, and Yoshua Bengio.
\newblock 2010.
\newblock Word representations: A simple and general method for semi-supervised
  learning.
\newblock In {\em Proceedings of the 48th Annual Meeting of the Association for
  Computational Linguistics}, pages 384--394, Uppsala, Sweden, July.
  Association for Computational Linguistics.

\bibitem[\protect\citename{Wan \bgroup et al.\egroup
  }2013]{wan2013regularization}
Li~Wan, Matthew Zeiler, Sixin Zhang, Yann~L Cun, and Rob Fergus.
\newblock 2013.
\newblock Regularization of neural networks using dropconnect.
\newblock In {\em Proceedings of the 30th International Conference on Machine
  Learning (ICML-13)}, pages 1058--1066.

\bibitem[\protect\citename{Wang and Manning}2012]{wang2012baselines}
Sida Wang and Christopher Manning.
\newblock 2012.
\newblock Baselines and bigrams: Simple, good sentiment and topic
  classification.
\newblock In {\em Proceedings of the 50th Annual Meeting of the Association for
  Computational Linguistics (Volume 2: Short Papers)}, pages 90--94, Jeju
  Island, Korea, July. Association for Computational Linguistics.

\bibitem[\protect\citename{Wang \bgroup et al.\egroup }2015]{genCNN}
Mingxuan Wang, Zhengdong Lu, Hang Li, Wenbin Jiang, and Qun Liu.
\newblock 2015.
\newblock gencnn: A convolutional architecture for word sequence prediction.
\newblock In {\em Proceedings of the 53rd Annual Meeting of the Association for
  Computational Linguistics and the 7th International Joint Conference on
  Natural Language Processing (Volume 1: Long Papers)}, pages 1567--1576,
  Beijing, China, July. Association for Computational Linguistics.

\bibitem[\protect\citename{Yin and Sch\"{u}tze}2015]{Yin2}
Wenpeng Yin and Hinrich Sch\"{u}tze.
\newblock 2015.
\newblock Multichannel variable-size convolution for sentence classification.
\newblock In {\em Proceedings of the Nineteenth Conference on Computational
  Natural Language Learning}, pages 204--214, Beijing, China, July. Association
  for Computational Linguistics.

\bibitem[\protect\citename{Yin \bgroup et al.\egroup }2016]{Yin}
Wenpeng Yin, Sebastian Ebert, and Hinrich Sch{\"u}tze.
\newblock 2016.
\newblock Attention-based convolutional neural network for machine
  comprehension.
\newblock {\em arXiv preprint arXiv:1602.04341}.

\bibitem[\protect\citename{Zeiler}2012]{zeiler2012adadelta}
Matthew~D Zeiler.
\newblock 2012.
\newblock Adadelta: an adaptive learning rate method.
\newblock {\em arXiv preprint arXiv:1212.5701}.

\bibitem[\protect\citename{Zhang \bgroup et al.\egroup }2015]{Zhang}
Xiang Zhang, Junbo Zhao, and Yann LeCun.
\newblock 2015.
\newblock Character-level convolutional networks for text classification.
\newblock In {\em Advances in Neural Information Processing Systems}, pages
  649--657.

\bibitem[\protect\citename{Zhang \bgroup et al.\egroup
  }2016]{zhang2016dependency}
Rui Zhang, Honglak Lee, and Dragomir~R. Radev.
\newblock 2016.
\newblock Dependency sensitive convolutional neural networks for modeling
  sentences and documents.
\newblock In {\em Proceedings of the 2016 Conference of the North American
  Chapter of the Association for Computational Linguistics: Human Language
  Technologies}, pages 1512--1521, San Diego, California, June. Association for
  Computational Linguistics.

\bibitem[\protect\citename{Zhao and Wu}2016]{zhao2016attention}
Zhiwei Zhao and Youzheng Wu.
\newblock 2016.
\newblock Attention-based convolutional neural networks for sentence
  classification.
\newblock {\em Interspeech 2016}, pages 705--709.

\end{thebibliography}
\bibliographystyle{eacl2017}
\newpage
\appendix
\section{Supplemental Material}
\subsection{Selection of $k$-grams}
\textbf{Dictionary based Selection.} The dictionary of positive and negative words that we used is available\footnote{https://github.com/jeffreybreen/twitter-sentiment-analysis-tutorial-201107/tree/master/data/opinion-lexicon-English}. 

\textbf{Parse Trees.} We used the Stanford parser (version 3.6, NLTK Python interface) to generate the parse tree for the sentences. All the $k$-grams having 3,4 and 5 words are added to the set selected using the above mentioned dictionary based set.

\textbf{POS-Tag.} We used the Stanford POS tagger to tag words in a sentence.

\subsection{Experiments Details}
\textbf{Variable length sentences.} The variable length sentences are handled by padding zeros. \\

\textbf{Training related hyper-parameters.}
\begin{table} [H]
          \begin{tabular}{|c|c|} 
     \hline  
     \textbf{Hyperparameter} & \textbf{Value} \\ 
     \hline \hline 
     Kernel Window Sizes ($k$-grams) & 3,4,5 \\
      \hline  
     Number of Kernels (for each $k$-gram) & 100\\ 
      \hline  
      Regularization & L1 \\
     \hline  
    Regularization parameter ($\lambda$) range & 1e-06 to 1e-08\\ 
     \hline  
    Dropout & 0.5\\ 
     \hline  
    Weights for Word2Vec and SentiWordNet Representations & 1, 10\\
    (Distance function computation) & \\ 
		\hline
      \end{tabular} 
      
    \caption{Hyperparameters} 
\end{table}
For the AdaDelta update rule, we used the same hyperparameters suggested in~\cite{zeiler2012adadelta}. 

\subsection{Datasets Details}
\begin{itemize}
    \item \textbf{Movie Review (MR)} : Movie review dataset contains 5331 positive and 5331 negative reviews introduced in \cite{pang2005}\footnote{https://www.cs.cornell.edu/people/pabo/movie-review-data/}. 
    \item \textbf{SUBJ} : The dataset contains 5000 subjective and 5000 objective sentences introduced in \cite{pang2004sentimental}\footnote{https://www.cs.cornell.edu/people/pabo/movie-review-data/}. The problem is to classify a given sentence as subjective or objective.
    \item \textbf{SST1} : Stanford Sentiment Treebank 1 is an extension to MR with the Train/Test/Dev splits. The dataset is a fine grained version with 5 labels(very positive, positive, neutral, negative and very negative) introduced in \cite{socher2013recursive} \footnote{http://nlp.stanford.edu/sentiment/}.
    \item \textbf{SST2} : Stanford Sentiment Treebank 2 is a subset to SST1 with neutral reviews removed, positive and very positive labeled as positive, negative and very negative labeled as negative.
    \item \textbf{IMDB} : Internet Movie DataBase is the large movie review dataset with 25000 train and 25000 test samples~\cite{maas-EtAl:2011:ACL-HLT2011}\footnote{http://ai.stanford.edu/~amaas/data/sentiment/}.
\end{itemize}
While the dataset \textbf{SST1} corresponds to a 5-class problems, rest of the datasets are binary classification problems.

\end{document}